\documentclass[onecolumn]{IEEEtran}

\usepackage{cite}
\usepackage[cmex10]{amsmath}

\usepackage{amsmath}
\usepackage{algorithm}
\usepackage{algorithmic}
\usepackage{hyperref}  
 
\newtheorem{instance}{Instance} 
\newtheorem{theorem}{Theorem}
\newtheorem{definition}{Definition}

\newcommand{\vioa}{\mathrm{vio1}} 
\newcommand{\viob}{\mathrm{vio2}} 
\newcommand{\vioc}{\mathrm{vio3}} 
\newcommand{\loc}{\mathrm{loc}}  
\begin{document}

\title{Analysis of Solution Quality of a Multiobjective Optimization-based Evolutionary Algorithm  for Knapsack Problem} 
\author{Jun He, Yong Wang  and Yuren Zhou
\thanks{This work was partially supported by EPSRC under Grant No. EP/I009809/1 (He), by NSFC under Grant Nos. 61170081, 61472143 (Zhou), 61273314 and by the Program for New Century Excellent Talents in University under Grant NCET-13-0596 (Wang).}   
\thanks{ Jun He is with Department of Computer Science,
Aberystwyth University, Aberystwyth, UK}
\thanks{ Yong Wang  is with School of Information Science and Engineering, Central South University, Changsha 410083, China}
\thanks{Yuren Zhou is with School of Advanced Computing,
Sun Yat-sen University,
Guangzhou, 510006,  China
}
}

\maketitle

\begin{abstract}
Multi-objective optimisation is regarded as one of  the most promising ways for dealing with constrained  optimisation problems in evolutionary optimisation.   This paper presents a theoretical investigation of  a  multi-objective optimisation evolutionary algorithm  for  solving the 0-1 knapsack problem. Two initialisation methods are considered in the algorithm:   local search initialisation  and greedy search initialisation.  Then  the solution quality  of the algorithm is analysed in terms of  the approximation ratio.  
\end{abstract} 

\section{Introduction}
  Consider the problem of maximizing an objective function,
\begin{equation}
\label{proConstrained}
\max_{\vec{x}}   f(\vec{x}) ,   
\quad  \mbox{subject to } g(x) \le 0.
\end{equation} 
The above constrained optimisation problem can be transferred into an unconstrained bi-objective optimisation problem. That is to optimize the original objective function plus  to minimize the constraint violation simultaneously: 
\begin{equation}
\left\{
\begin{array}{lll}
 \max_{\vec{x}} f(\vec{x}),\\
 \min_{\vec{x}} v(\vec{x}),
\end{array} 
\right.
\end{equation}
where $v(\vec{x})$ is the degree of constraint violation, given by
\begin{align}  v(\vec{x})=
\left\{
 \begin{array}{lll}
0, &\mbox{if }g(\vec{x}) \le 0, \\
 g(\vec{x}), &\mbox{otherwsie.}
\end{array} 
\right.
\end{align}

The use of multi-objectives for single-objective optimisation problems could be traced back to 1990s~\cite{louis1993pareto}. This   methodology has been termed multiobjectivisation \cite{knowles2001reducing}. Using multiobjectivization sometimes may help the search more efficient as shown in  \cite{jensen2005helper,neumann2006minimum,neumann2007expected,friedrich2010approximating}.

According to the survey~\cite{segura2013using}, multi-objective optimisation is regarded as one of  the most promising ways for dealing with constrained     optimisation problems in evolutionary optimisation. A constrained optimisation problem is often transformed into a bi-objective optimisation problem, in which the first objective is the original objective function and the second objective is the degree of constraint violation~\cite{zhou2003multi,cai2006multiobjective,wang2007multiobjective,wang2008adaptive}.   After this transformation, Pareto dominance is frequently employed to compare individuals. Currently the research in this area is very active \cite{segura2013using}. For example,  a self-adaptive selection method is proposed recently in \cite{jiao2013novel}, which aims to exploit both non-dominated solutions with low constraint violations and feasible solutions with low objective function values.    
Multi-objective optimisation is combined with differential evolution in \cite{wang2012combining} and  an infeasible solution replacement mechanism is proposed.   A dynamic hybrid framework is presented in \cite{wang2012dynamic}, where the global and local search models are implemented dynamically according to the feasibility proportion of the   population.

This paper aims at analysing the solution quality of  evolutionary algorithms (EAs)  in terms of the approximation ratio. It is not intended to demonstrate that EAs are able to compete with problem-specific  approximation algorithms, since this is unlikely in most cases. Nevertheless,  it is still necessary and important  to understand the solution quality of EAs, so  the  EAs with arbitrarily bad solution quality  could be avoided in applications.  The analysis of the approximation performance of EAs has attracted a lot of interests  in recent  years~\cite{friedrich2009analysis,oliveto2009analysis,lai2014performance}.

This paper  investigate  an existing multiobjective optimization-based   EA~\cite{cai2006multiobjective} (MOEA) for solving constrained optimisation problems.   The   MOEA   originally is designed for continuous optimization. Here it is adapted for solving  the 0-1 knapsack problem.   Although experiment results show its performance is good, no theoretical analysis exists for this MOEA \cite{cai2006multiobjective}. This motivates our    rigorous  analysis.

The remainder of the paper is organized as follows.  The 0-1 knapsack problem and approximation ratio are introduced in Section \ref{secProblem}. The MOEA with the local search initialisation is analysed in Section~\ref{secMOEA} .    Section~\ref{secHybrid} is devoted to the analysis of the MOEA with the greedy  search initialisation.  Section~\ref{secConclusions} concludes the article.

\section{0-1 Knapsack Problem and Approximation Ratio of Solutions}
\label{secProblem}
Given an instance of the 0-1 knapsack problem  with a set of weights
$w_i$, values $v_i$, and capacity $W$ of a knapsack, the task is to find a binary string $\vec{x}_{\max}$ so as to maximize the objective function,
\begin{equation} 
\max_{\vec{x}}   f(\vec{x})= \sum^n_{i=1} v_i x_i,   
\quad  \mbox{subject to } \sum^n_{i=1} w_i x_i \le W,
\end{equation}
where  $\vec{x}=(x_1\cdots x_n)$ is a binary string. $
x_i =   1$ if  item $i$ is
selected in the knapsack; otherwise $x_i=
0$. 

A \emph{feasible solution} is a knapsack represented by an $\vec{x}$ which satisfies the constraint, that is $\sum^n_{i=1} w_i x_i \le W$. An \emph{infeasible} one is an $\vec{x}$ that violates the constraint.  The string $(0 \ldots  0)$ represents a null knapsack. 
 Without loss of generality, assume that   a feasible solution always exists and    $n$ is large.

There exist  well-known  approximation algorithms for the 0-1 knapsack problem~\cite{martello1990knapsack,kellerer2004knapsack}. Probably the simplest one is the greedy search~\cite{martello1990knapsack} whose worst-case approximation performance ratio equals to 1/2 and time complexity is $O(n)$ plus $O(n \log n)$ for the initial sorting. A polynomial-time approximation scheme has been introduced in \cite{martello1990knapsack} whose worse-case performance is $k/(1+k)$ given an integer parameter $k$ and its time complexity is $O(n^{k+1})$. Furthermore,   a fully-polynomial-time approximation scheme is well-known \cite{martello1990knapsack} whose time complexity is $O(n/\epsilon^2)$ plus $O(n \log n)$ for the initial sorting   given  a parameter $\epsilon>0$.

In evolutionary optimisation, the 0-1 knapsack problem has been  taken  as a benchmark  in computer experiments \cite{michalewicz1994genetic,michalewicz1996genetic}   for evaluating the performance of various constraint-handling techniques.   
It is also  one of  favourite problems used in the theoretical study of EAs \cite{kumar2006analysis,zhou2007runtime}.

In order to assess the  solution quality of an EA, an evolutionary approximation algorithm  is defined  as below. It follows the definition of conventional $\alpha$-approximation algorithms  \cite[Definition 1]{williamson2011design}.
\begin{definition}\label{alphaApproxDefn}
An EA is an \emph{$\alpha$-approximation algorithm} for a constrained optimisation problem if for all instances of the problem, the EA can produce a feasible solution  in   polynomial running time, whose objective function value   is within a factor of $\alpha$ of that of an optimal solution. The \emph{running time} of an EA is the expected number of function evaluations.
\end{definition}

In a maximisation problem (assume $f(\vec{x}_{\max})>0$), a feasible solution $\vec{x}$ is called to have  an \emph{$\alpha$-approximation ratio}  if it satisfies 
\begin{align}
\frac{f(\vec{x})}{f(\vec{x}_{\max})} \ge \alpha.  
\end{align}

 In order to prove that an EA is  not an $\alpha$-approximation algorithm, it is sufficient to  show that an EA needs exponential running time to obtain a feasible solution  with  an $\alpha$-approximation ratio in  one instance of the problem.

\section{Analysis of  MOEA with Local Search Initialisation}
\label{secMOEA}

The 0-1 knapsack problem can be transformed into a bi-objective optimisation problem,  that is to maximize the objective function $ f(\vec{x})$ and to minimize the constrain violation $v(\vec{x})$, 
where  $v(\mathbf{x})$ is defined by
 \begin{align}
\label{equViolation}
v(\mathbf{x})=
\left\{
\begin{array}{llll}
0; &\mbox{if $\mathbf{x}$ is feasible},\\
\sum^n_{i=1} w_i -W, &\mbox{otherwise}.
\end{array}
\right.
\end{align}

Although a constrained optimisation problems can be converted into a bi-objective optimisation problem, there exists an essential difference between it and general multi-objective optimisation problems~\cite{cai2006multiobjective}. The target of   general multi-objective optimisation is to obtain a final population with a diverse non-dominated individuals uniformly distributed on the Pareto front. However in the bi-objective optimisation problem derived from contained optimisation, the target is to obtain the optimal feasible solution of  the original  constrained optimisation problem. Consequently, there is no need to care about the uniform distribution of the resulting solutions on the Pareto front.

The MOEA adopted in this section is a variant of an existing MOEA  proposed in \cite{cai2006multiobjective}. The fundamental idea in this MOEA is that non-dominated individuals in a children population
are chosen and replace dominated individuals of the parent population.  
The algorithm,  based on  Model 1 of \cite{cai2006multiobjective},  is described in Algorithm~\ref{alg1} for solving the 0-1 knapsack problem.  
\begin{algorithm}[ht]
\caption{MOEA  \cite{cai2006multiobjective}}
\label{alg1}
\begin{algorithmic} [1] 
 \STATE initialize  population $\Phi_0$;
\FOR{$t=0, 1,2, \cdots$}
\STATE    perform bitwise mutation and generate  a children  population $\Phi_{t.a}$ with $N$ individuals;
\STATE evaluate  the values of $f(\vec{x})$ and $v(\vec{x})$;
\STATE choose the non-dominated individuals  from population $\Phi_{t.a}$ and assume there are $k$ non-dominated individuals, denoted as $ \{ \vec{x}_1, \cdots, \vec{x}_k\}$;  
\STATE set an intermediate population $\Phi_{t.b}\leftarrow \Phi_t$;
\FOR{$i=1, \cdots, k$}
\STATE  let $m$ be the number of individuals in $\Phi_{t.b}$ which are dominated by $\vec{x}_i$;
\IF{$m=0$}
\STATE do nothing;
\ELSIF{$m=1$} 
\STATE the corresponding dominated individual is replaced by $\vec{x}_i$;
\ELSE 
\IF{the dominated individuals are feasible}
\STATE the individual with the smallest objective function value   is replaced by $\vec{x}_i$;
\ELSE
\STATE one of the dominated individuals is randomly chosen and  replaced by $\vec{x}_i$;
\ENDIF
\ENDIF
\ENDFOR
\STATE   set the next generation population $\Phi_{t+1} \leftarrow \Phi_{t.b}$. 
\ENDFOR 
\end{algorithmic}
\end{algorithm}

 It should be pointed out that the running time of EAs is dependent on initialisation. Two initialisation methods are considered in the paper: initialisation by the local search and by the greedy search.  In this section, we investigate the first one: the local search initialisation, described in Algorithm~\ref{algLocal}. This initialisation  does not only produce both feasible solutions (local optima), but also infeasible solutions. Bitwise mutation  flips each bit of a binary string  with   probability $\frac{1}{n}$. Population size $N$ is set to a large constant and for the sake of analysis, assume that $N/4$ is an integer.

\begin{algorithm}
\caption{Local Search Initialisation}
\label{algLocal}
\begin{algorithmic} [1]
\STATE set $\vec{x}=(0\cdots0)$;
 \WHILE{$\vec{x}$ is feasible}
 \STATE flip one 0-valued bit of $\vec{x}$ into 1-valued, denote it by $\vec{y}$;
 \IF{$\vec{y}$ is feasible} 
 \STATE let $\vec{x} \leftarrow\vec{y}$;
 \ELSE 
 \STATE let $\vec{x} \leftarrow\vec{y}$ with probability 1/2;
 \ENDIF
 \ENDWHILE
 \STATE repeat the above steps until $N$ individuals are produced.
\end{algorithmic} 
\end{algorithm}

We show that the solution quality of the MOEA with the local search initialisation might be  arbitrarily bad using the following   instance of the 0-1 knapsack problem.

\begin{instance}   
In the following table, $H$, $I$ and $J$ represent  index sets.  
For the sake of simplicity, assume that $\frac{n}{2}$ and $\frac{\alpha n}{2}$ are integers. Fixing a  constant $\alpha \in (0,1)$, choose $n$ an enough large integer so that   $n >  \frac{2}{\alpha}$ and $\frac{2}{\alpha} >  \frac{n \alpha^{2 \ln n} }{2}$. 

\begin{table}[ht]
\begin{center}
\begin{tabular}{c|c|c|c}
\hline
&$H$& $I$ & $J$ \\
\hline
$i$ & $1$ & $ 2,  \cdots,  \frac{n}{2} +1  $ & $ \frac{n}{2}  +2, \cdots,  n$ \\
\hline
 $v_i$ &  $n$ &$1$     & $\alpha^{\ln n} $  \\
$w_i$ &  $n$  &  $ \frac{2}{\alpha}  $  &$ \alpha^{2\ln n}$     \\
\hline
$W$ &\multicolumn{3}{c}{$  n  $}\\
\hline
\end{tabular}
\end{center}
\caption{Instance \ref{insMulti}}\label{insMulti}
 \end{table}
\end{instance}
 
Let $\vec{x}_{\max}$  represent the global optimum  such that $x_1= 1$ and other bits $x_i=0$. The global optimum is unique. Its objective function value is
\begin{align}
f(\vec{x}_{\max})=n .
\end{align}

Let  $\vec{x}_\loc$ represent a local optimum\footnote{A feasible solution $\vec{x}$ is called a \emph{local optimum} if  $f(\vec{y})< f(\vec{x})$ for any feasible solution $\vec{y}$ within  Hamming distance $d(\vec{x},\vec{y})=  1$. }  such that $\frac{\alpha n}{2}$ bits $x_i=1$ (where $i \in I$)   and other bits $\vec{x}_i=0$. Its objective function value is
\begin{align}
f(\vec{x}_{\loc})= \frac{\alpha n}{2} .
\end{align}
 $f(\vec{x}_{\loc})$ is the second largest objective function  value among feasible solutions. 
The number of such local optima $\vec{x}_{\loc}$ is experiential in  $n$,
\begin{align}
\binom{\frac{n}{2} }{\frac{\alpha n}{2}} \ge \left(\frac{1}{\alpha}\right)^{\frac{\alpha n}{2}}.
\end{align}
 
Let $\vec{x}_{\vioa}$  denote an infeasible solution   such that $x_i =1$  for $\frac{\alpha n}{2}$ indexes $i \in I$, one  $i \in J$, and $x_i=0$ for other $i$.  Its objective function value and violation value are
\begin{align}
f(\vec{x}_{\vioa})= f(\vec{x}_{\loc})+\alpha^{\ln n}, \quad v(\vec{x}_{\vioa})=\alpha^{2\ln n}.
\end{align}  The number of such infeasible solutions $\vec{x}_{\vioa}$ is    exponential in  $n$,
\begin{align}
 \binom{\frac{n}{2} }{\frac{ \alpha n}{2}} \frac{n-1}{2} \ge \frac{n-1}{2} \left(\frac{1}{\alpha}\right)^{\frac{\alpha n}{2}}.
\end{align}

Let $\vec{x}_{\viob}$  denote an infeasible solution such that $x_1=1$,  $x_i=1$ for one $i \in J$, and $x_i=0$ for any other $i$. Its objective function value and violation value are
\begin{align}
f(\vec{x}_{\viob})= f(\vec{x}_{\max})+\alpha^{\ln n}, \quad v(\vec{x}_{\viob})=\alpha^{2\ln n}.
\end{align} 
 
It is easy to verify that the degree of constraint violation $v(\vec{x}_{\vioa})$ and $v(\vec{x}_{\viob})$ $(=\alpha^{\ln n})$ is the minimum among all infeasible solutions.
 
Instance 1 is hard since Hamming distance between a local optimum $\vec{x}_{\loc}$ and the unique global optimum $\vec{x}_{\max}$ is large $\ge \alpha n/2$ and the the number of local optima is exponential in $n$. Since the selection used in the MOEA is that non-dominated individuals in a children population
are chosen and replace dominated individuals of the parent population, it prevents individuals moving from the 2nd best fitness level to the best fitness level. 
Thus the EA needs exponential time to leave the absorbing basin of the local optima.

\begin{theorem}
\label{thmMulti}
For Instance 1 and any constant $\alpha \in (0,1)$,  the MOEA with the local search initialisation needs    $\Omega(n^{ \frac{\alpha n}{2}})  $ running time to find an   $\alpha$-approximation solution in the worst case.
\end{theorem}

\begin{IEEEproof}
 After the initialisation, individuals generated by the local search may include $\vec{x}_{\max}$,   local optima $\vec{x}_\loc$ and infeasible solutions $\vec{x}$ with  Hamming distance $H(\vec{x}, \vec{x}_{\max})=1$ or $H(\vec{x}, \vec{x}_\loc)=1$.
The worst case is that after   initialisation, population $\Phi_0$ is  composed of $N$    local optima $\vec{x}_{\loc}$ and infeasible solutions  $\vec{x}_{\vioa}$. 
Notice that the number of   local optima $\vec{x}_{\loc}$  and   infeasible solutions $\vec{x}_{\vioa}$  is  exponential in  $n$.  The individuals in $\Phi_0$ may be chosen to be  different.

Assume that in the $t$th generation, population $\Phi_t$ is  composed of $N$   local optima $\vec{x}_{\loc}$ and  infeasible solutions  $\vec{x}_{\vioa}$.  The approximation ratio  between $f(\vec{x}_{\loc})$ and $f(\vec{x}_{\max})$  is 
\begin{align}
\frac{f(\vec{x}_{\loc})}{f(\vec{x}_{\max})} =\frac{\alpha}{2} < \alpha.
\end{align} 
Since $\alpha$ could be any constant,  the above approximation ratio   could arbitrarily bad. 

As we know that $\vec{x}_{\max}$ is the unique solution  satisfying  $f(\vec{x}_{\max}) > f(\vec{x}_{\loc})$,  
it is sufficient to prove that  the EA needs exponential  running time to generate  $\vec{x}_{\max}$.  

First we consider the event of mutating $\vec{x}_{\loc}$ or $\vec{x}_{\vioa}$ into a child $\vec{y}$.   The event can be decomposed into the following mutually exclusive and exhaustive  sub-events.

\begin{enumerate}
\item  $\vec{y}$ is a feasible solution such that $f(\vec{y}) < f(\vec{x}_{\loc})$. 

Obviously $\vec{y}$ will not dominate $\vec{x}_{\loc}$. At the same time, $\vec{y}$ will not dominate $\vec{x}_{\vioa}$ since $f(\vec{y}) < f(\vec{x}_{\loc}) < f(\vec{x}_{\vioa})$. Thus $\vec{y}$ will not dominate any individuals in $\Phi_t$ and cannot be selected into the next generation population.

\item   $\vec{y}$ is a feasible solution such that $f(\vec{y})= f(\vec{x}_{\loc})$, that is, $\vec{y}$ is a $\vec{x}_{\loc}$.   

\item $\vec{y}$ is a feasible solution such that $f(\vec{y})> f(\vec{x}_{\loc})$, that is,   $\vec{y}$ is the global optimum $\vec{x}_{\max}$.

 In this case,   all 1-valued bits $x_i$ (where $i \in I$) must be flipped from 1 to 0.
The probability of the event happening is at most    
\begin{align}
O\left( \frac{1}{n} \right)^{\frac{\alpha n}{2}}  .
\end{align}

\item  $\vec{y}$ is an infeasible solution such that $v(\vec{y}) > v(\vec{x}_{\vioa})$. 

In this case, $\vec{y}$ will not dominate $\vec{x}_{\loc}$ and $\vec{x}_{\vioa}$, so it cannot be selected into the next generation population.

\item  $\vec{y}$ is an infeasible solution such that $v(\vec{y}) =v(\vec{x}_{\vioa})$, that implies, 
\begin{enumerate}
\item either $\vec{y}$  is   $\vec{x}_{\vioa}$;
\item or $\vec{y}$  is an infeasible solution $\vec{x}_{\viob}$.

In the second case,  all 1-valued bits $x_i$ (where $i \in I$) must be flipped from 1 to 0.
The probability of the event happening is   
\begin{align}
O\left( \frac{1}{n} \right)^{\frac{\alpha n}{2}}  .
\end{align}
\end{enumerate}

\end{enumerate}

From the above analysis, we observe that only when either $\vec{x}_{\max}$  or an infeasible solution $\vec{x}_{\viob}$ is generated via mutation,   the population status could be changed. Otherwise the population is still composed of $N$ solutions $\vec{x}_{\loc}$ and $\vec{x}_{\vioa}$. The probability of the former event happening is   
$
O\left( \frac{1}{n} \right)^{\frac{\alpha n}{2}}  .
$

Next we analyse the role of using a population. Consider  the event that a population includes either $\vec{x}_{\max}$    or an infeasible solution $\vec{x}_{\viob}$ is generated via mutation.  Since   $N$ parents are selected and mutated independently,   the probability of the event happening is  
$
  N O \left(  {n}^{\frac{2}{\alpha n}} \right).
$
Thus the expected number of generations for the EA to reach $\vec{x}_{\max}$ is 
$ \frac{1}{N} \Omega \left( n^{\frac{\alpha n}{2}} \right).
$
Since there are $N$ fitness evaluations at each generation, the expected number of fitness evaluations is 
$    \Omega \left( n^{\frac{\alpha n}{2}}\right).
$
The required conclusion is then proven. 
\end{IEEEproof}

\section{Analysis of  MOEA with  Greedy Search Initialisation}
\label{secHybrid}
In order to produce a good quality solution  with a guaranteed approximation ratio, a natural idea is to combine an EA with an approximation algorithm: first we apply an approximation algorithm to producing approximation solutions as the initial population, and then   apply the MOEA  to searching  a global optimum. In this section, we consider a $\frac{1}{2}$-approximation algorithm to implement the initialisation. It a variant of the greedy search~\cite[Section 2.4]{martello1990knapsack}, described in Algorithm~\ref{algGreedy}. Notice that the initialisation does not only produce feasible solutions (local optima), but also infeasible solutions.

\begin{algorithm}
\caption{Greedy Search Initialisation}
\label{algGreedy}
\begin{algorithmic} [1]
\STATE   sort all the items via their values  so that $p_1\ge \cdots \ge p_n$;
\STATE
then greedily add the items in the above order to the knapsack as long as adding an item to the knapsack does not exceeding the capacity of the knapsack. Denote the solution by $\vec{x}_{a}$;

 \STATE resort all the items via the ratio of their values to their corresponding weights so that $\frac{p_1}{w_1} \ge \cdots \ge  \frac{p_n}{w_n}$;
 \STATE
 Then greedily add the items in the above order to the knapsack as long as adding an item to the knapsack does not exceeding the capacity of the knapsack. Denote the solution by $\vec{x}_{b}$;
 
 \STATE put  $\vec{x}_{a}$ and $\vec{x}_{b}$ into the initial population;
 
 \STATE repeat the above procedure until $\frac{N}{2}$ individuals are produced;
 
 \STATE for each of these $\frac{N}{2}$ individuals, add one item and then $\frac{N}{2}$ infeasible solutions are produced. 
\end{algorithmic} 
\end{algorithm}

Using the greedy search initialisation, we may find the global optimal solution of Instance 1  during the initialisation phase. Furthermore, since the greedy search is a $1/2$-approximation algorithm for the 0-1 knapsack problem,  the MOEA with the greedy  search initialisation   is an evolutionary  $1/2$-approximation algorithm too.  The advantage of using an EA is the ability   to obtain the global optimum due to the use of bitwise mutation.

In the following we answer the question:  can the MOEA with the greedy search initialisation  find a solution with  the approximation ratio better than 1/2? Through   analysing the instance described below, we obtain a negative answer. 

\begin{instance} 
In the following table, $H$, $I$,  $J$ and $K$  represent index sets.   For the sake of simplicity, assume that ${n}/{4}$ is an integer. 
\begin{table}[ht]
\begin{center}
\begin{tabular}{c|c|c|c |c}
\hline
     & $H$         & $I$ & $J$ &$K$ \\
     \hline
$i$ & $  1,  2 $ &   $ 3, \cdots,  \frac{n}{4} $ &$\frac{n}{4} +1, \cdots, \frac{n}{2} $ & $\frac{n}{2}+1, \cdots, n$ \\
\hline
 $v_i$   & $n$      &$ n+2$      & $n^{-3} $  & $n^{-3}$   \\
$w_i$ &   $n$   &  $n+1$     & $n^{-4} $   & $\frac{1}{4(1+n^{-4})}$ \\
\hline
$W$ &\multicolumn{4}{c}{$ 2n  $}\\
\hline
\end{tabular}
\end{center}
\caption{Instance \ref{insMulti2}}\label{insMulti2}
 \end{table}
\end{instance}
 
Let $\vec{x}_{\max}$ represent  the unique global optimum such that $x_i=1$ for any  $i\in H$ and $x_i=0$ for any other  $i$.  Its objective function value is
\begin{align}
f(\vec{x}_{\max})=2n .
\end{align}

Let $\vec{x}_{\loc}$ represent  a local optimum such that $x_i=1$ for   one  $i \in I  $, any   $i \in J$ and $\frac{n}{4}-2$ indexes $i \in K$;  $x_i=0$ for all other  $i$. 
Its objective function value is
\begin{align}
f(\vec{x}_{\loc})=  n+2+\left(\frac{n}{2}-2\right)n^{-3} .
\end{align} $f(\vec{x}_{\loc})$ is the second largest objective function  value among feasible solutions.

Let $\vec{x}_{\vioa}$  represent  an infeasible solution  such that $x_i=1$ for   one  $i \in I  $, any  $i \in J$ and $\frac{n}{4}-1$  indexes $i \in K$;  $x_i=0$ for all other   $i$.  Its objective function value and violation value are
\begin{align}
f(\vec{x}_{\vioa})= f(\vec{x}_{\loc}) +n^{-3}, \quad v(\vec{x}_{\vioa})= \frac{1}{4(1+n^{-4})} .
\end{align} 

Let $\vec{x}_{\viob}$  represent  an infeasible solution such that $x_i=1$ for any   $i\in H$ and   one   $i \in K$, and $x_i=0$ for any other  $i$.  Its objective function value and violation value satisfy
\begin{align}
 f(\vec{x}_{\viob})=  f(\vec{x}_{\max}) +n^{-3}, \quad   v(\vec{x}_{\viob})=\frac{1}{4(1+n^{-4})}.
\end{align}   

Let $\vec{x}_{\vioc}$  represent  an infeasible solution such that $x_i=1$ for any  $i\in H$ and at least one   $i \in J$, and $x_i=0$ for any other  $i$.  Its objective function value and violation value satisfy
\begin{align}
f(\vec{x}_{\max})  < f(\vec{x}_{\vioc})\le  f(\vec{x}_{\max}) +\frac{n^{-2}}{4}, \quad  0< v(\vec{x}_{\vioc})<\frac{n^{-3}}{4}.
\end{align} 
  
Instance 2 is a hard problem to EAs using bitwise mutation since Hamming distance between a local optimum $\vec{x}_{\loc}$ and the unique global optimum $\vec{x}_{\max}$ is large. Another trouble is the number of local optima which is exponential in $n$. 
Thus it is difficult for a population  to leave the absorbing basin of the local optima.

\begin{theorem}
For Instance 2, the MOEA with the greedy search initialisation can find a  $\frac{1}{2}$-approximation solution after initialisation.
But in the worst case, it needs $\Omega( n^{\frac{n}{4}})$  running time to find   a $(\frac{1}{2}+\frac{1}{n}+\frac{1}{2 n^{3}})$-approximation solution.
\end{theorem}

\begin{IEEEproof}
The first conclusion is trivial  due to the use of the greedy search.
 After the initialisation, individuals generated by the greedy search are   local optima $\vec{x}_\loc$ and infeasible solutions $\vec{x}$ with  Hamming distance $H(\vec{x}, \vec{x}_\loc)=1$. The local optimum $\vec{x}_{\loc}$ has an approximation ratio  given by
\begin{equation}
\begin{split}
\frac{f(\vec{x}_{\loc})}{f(\vec{x}_{\max})} 
=\frac{n+2+\left(\frac{n}{2}-2\right)n^{-3} }{2n} \in \left(\frac{1}{2},  \frac{1}{2}+\frac{1}{n}+\frac{1}{2 n^{3}}\right).
\end{split}
\end{equation}

The  proof of the second conclusion is similar to that of Theorem 1. 
The worst case is that after   initialisation, population $\Phi_0$ is  composed of $N$     local optima $\vec{x}_{\loc}$  and infeasible solutions  $\vec{x}_{\vioa}$. 
Notice that the number of   local optima $\vec{x}_{\loc}$  and   infeasible solutions $\vec{x}_{\vioa}$  is  exponential in  $n$.  The individuals in $\Phi_0$ could be chosen to be different.

Assume that in the $t$th generation, population $\Phi_t$ is  composed of $N$  different local optima $\vec{x}_{\loc}$ and infeasible solutions  $\vec{x}_{\vioa}$.   Let $\mathbf{y}$ be a child mutated from a parent $\mathbf{x}$.  The event can be decomposed into the following mutually exclusive and exhaustive  sub-events.

\begin{enumerate}
\item $\mathbf{y}$ is feasible such that $f(\mathbf{y}) < f(\mathbf{x}_{\loc})$. 

According to the selection based on the Pareto-dominance, $\mathbf{y}$ will not be selected to the next generation population.

\item $\mathbf{y}$ is feasible such that $f(\mathbf{y}) = f(\mathbf{x}_{\loc})$, that is,  $\mathbf{y}$ is  a local optimum $\vec{x}_{\loc}$.

\item $\mathbf{y}$ is feasible such that $f(\mathbf{y})> f(\mathbf{x}_{\loc})$, that is, $\vec{y}$ is $\vec{x}_{\max}$.  

In this case, any 1-valued bit $x_i$ where $i \in I \cup J \cup  K$ must be flipped from 1 to 0, and $x_1$ and $x_2$ must be flipped from 0 to 1. Thus at least $\frac{n}{4}$ bits must be flipped.  The probability of this event happening is at most $O(n^{-\frac{n}{4}})$.   

 \item $\mathbf{y}$ is infeasible such that $v(\mathbf{x})=\frac{1}{4(1+n^{-4})}$, that is
\begin{itemize}
\item   either   $\mathbf{y}$ is  $\vec{x}_{\vioa}$;

 \item or $\mathbf{y}$ is  $\vec{x}_{\viob}$.
 
 In the second case, except one 1-valued bit $x_i$ where $i \in K$, any other 1-valued bit $x_i$  where $i \in I \cup J \cup  K$ must be flipped from 1 to 0, and $x_1$ and $x_2$ must be flipped from 0 to 1. Thus at least $\frac{n}{4}$  bits must be flipped.  The probability of this event happening is at most $O(n^{-\frac{n}{4}})$.   
\end{itemize}

 \item $\mathbf{y}$ is infeasible such that $v(\mathbf{x})<\frac{1}{4(1+n^{-4})}$,  that means,   $\mathbf{y}$ is  $\vec{x}_{\vioc}$. 
 
 In this case,   any 1-valued bit $x_i$ where $i \in I \cup  K$ must be flipped from 1 to 0, and $x_1$ and $x_2$ must be flipped from 0 to 1. Thus at least $\frac{n}{4}$ bits must be flipped.  The probability of this event happening is at most $O(n^{-\frac{n}{4}})$.

 \item $\mathbf{y}$ is infeasible such that $v(\mathbf{x}) >\frac{1}{4(1+n^{-4})}$. $\mathbf{y}$ will not be selected  since it is dominated by $\vec{x}_{\vioa}$.
\end{enumerate}

From the above analysis, we observe that the probability of generating a non-$\vec{x}_{\loc}$ and non-$\vec{x}_{\vioa}$ child and   selecting it to the next generation population is small, that is $O(n^{-\frac{n}{4}})$.

Next we analyse the role of using a population. Consider the event that the next generation  population includes  a non-$\vec{x}_{\loc}$ or non-$\vec{x}_{\vioa}$ child.  Since   $N$ parents are mutated independently,   the probability of the event happening is at most $N O(n^{-\frac{n}{4}}).$ This implies  that the expected number of generations for the EA to reach $\vec{x}_{\max}$ is at least 
$  \frac{1}{N} \Omega(n^{ \frac{n}{4}}).$ 
Since there are $N$ fitness evaluations at each generation, the expected number of fitness evaluations is 
  $\Omega(  n^{\frac{n}{4}})$.
The required conclusion is   proven.
\end{IEEEproof}

The above theorem shows that the MOEA with the greedy search initialisation   finds   a $(\frac{1}{2}+\frac{1}{n}+\frac{1}{2 n^{3}})$-approximation solution in  $\Omega( n^{\frac{n}{4}})$  running time. As $n \to +\infty$, the approximation ratio goes towards $1/2$. In other words,  the MOEA doesn't   substantially improve the solution quality since the greedy search already produces a 1/2 approximation solution during   initialisation.

\section{Conclusions}
\label{secConclusions}
This paper has assessed the solution quality of  an existing MOEA~\cite{cai2006multiobjective}  for solving the 0-1 knapsack problem. The solution quality of an  EA is measured in terms of the   approximation ratio. Two different initialization methods are analysed in the MOEA: local search initialisation and greedy search initialisation. 
 
 When the initial population is produced by the local search, the solution quality of the  MOEA  might be arbitrarily bad in some instance. That is, given any constant $\alpha \in (0,1)$, the  MOEA   needs    $\Omega(n^{ \frac{\alpha n}{2}})  $ running time to find an   $\alpha$-approximation solution in the worst case in some instance. 

When the initial population is produced by the greedy search, the MOEA may guarantee  a 1/2-approximation solution within polynomial time. However, this improvement is caused by the use of the greedy search, rather than the MOEA itself. In some instance, the MOEA with the greedy search initialisation  needs $\Omega( n^{\frac{n}{4}})$  running time to find   a $(\frac{1}{2}+\frac{1}{n}+\frac{1}{2 n^{3}})$-approximation solution. In other words,  the MOEA doesn't   substantially improve the solution quality comparing with the greedy search.

Other types of initialisation, such as random initialisation, are not considered in the current paper. It is left for future work.

\end{document}